\documentclass{article} %
\usepackage[preprint]{sarp_preprint}

\usepackage{latexsym}
\usepackage[T1]{fontenc}
\usepackage[utf8]{inputenc}
\usepackage{microtype}
\usepackage{inconsolata}
\usepackage{xurl}
\usepackage{hyperref}
\usepackage{xcolor}
\definecolor{darkblue}{rgb}{0, 0, 0.5}
\hypersetup{colorlinks=true, citecolor=darkblue, linkcolor=darkblue, urlcolor=darkblue,
  pdftitle={Behavioral Compositionality of Sparse Autoencoder Representations}, pdfauthor={Haotian Ma, Ziye Mei, Frederic Sala}}

\usepackage{amsmath}
\usepackage{amssymb}
\usepackage{amsthm}
\usepackage{mathtools}

\usepackage{graphicx}
\usepackage{wrapfig}
\usepackage{booktabs}
\usepackage{array}
\usepackage{placeins}
\usepackage{needspace}

\newtheorem{theorem}{Theorem}

\newtheorem{corollary}{Corollary}

\DeclareMathOperator*{\argmin}{arg\,min}
\newcommand{\SAR}{\mathrm{SAR}}

\newcommand{\ind}{\mathbf{1}}

\title{Behavioral Compositionality of\\Sparse Autoencoder Representations}

\author{Haotian Ma \quad Ziye Mei \quad Frederic Sala \\
{\mdseries University of Wisconsin--Madison}}

\begin{document}
\maketitle

\begin{abstract}

Sparse autoencoders (SAEs) provide features that can be combined to explain and steer language model behavior. Understanding these combinations requires knowing how reliably individual feature effects predict their joint impact on the model. We study this relationship by retaining or removing subsets of SAE feature contributions and measuring the resulting behavioral responses.

We introduce SAE Additivity Residuals (SAR) to quantify how much behavioral variation is explained by feature interactions beyond a chosen order. We prove that, under independent feature subset sampling, SAR measures the normalized energy of these interactions and bounds errors in ranking feature combinations. Across 18 model and behavior settings, learned SAE coordinates have lower first order residuals than matched rotations of the same activation subspace. This shows that SAE coordinates support comparatively simple descriptions of joint feature effects. Pairwise modeling further improves the ranking of held out feature combinations in all 18 settings.

We also demonstrate the practical value of modeling these interactions. Across five challenging benchmarks, pairwise models select feature combinations that give higher probability to the correct answer and preserve the original behavioral response more closely than first order models. In MMLU-Pro steering, pairwise modeling improves rank fidelity from 0.5874 to 0.7790 and reduces ordering error from 0.1444 to 0.0606. These results establish behavioral compositionality as a measurable property of SAE representations and show how modeling feature interactions improves the analysis and selection of feature combinations.
Code is available at \href{https://github.com/OpenSAE/SAE-Behavioral-Compositionality}{\texttt{\mbox{https://github.com/}\allowbreak \mbox{OpenSAE/}\allowbreak \mbox{SAE-Behavioral-Compositionality}}}.
\end{abstract}

\section{Introduction}
\label{sec:introduction}

Sparse autoencoders (SAEs) decompose model activations into sparse, linearly
decoded features. These coordinates are used to inspect computation, build
multi-feature explanations, and manipulate behavior. This raises a basic
question. If the effects of individual features are known, can their joint
effect be predicted by summing them? A linear SAE decoder does not answer this
because subsequent computation can make feature effects reinforce, suppress, or
depend on one another.

This matters when SAE features are combined into circuits, joint explanations,
or interventions
\citep{marks2025sparse,ameisen2025circuit,ayonrinde2024interpretability}.
Existing SAE evaluations characterize reconstruction, sparsity, and
interpretability
\citep{karvonen2025saebench,engels2024dark,chanin2024absorption,engels2025notall},
but do not tell us when independently characterized features can be composed
into faithful joint explanations or interventions.

{\looseness=-1
To address this gap, we introduce a framework for measuring behavioral
compositionality in SAE coordinates. For each prompt, we retain or remove subsets of selected SAE decoded
contributions while keeping all other activation content fixed, then measure
the resulting downstream response over feature subsets. If individual
feature effects compose faithfully, this response should be well described at
first order. Departures from that approximation reveal interaction, and adding
successively higher interaction orders shows how that missing structure is
organized. This lets us compare learned SAE coordinates with matched rotations of the same
selected activation subspace, testing whether compositionality under these
interventions depends on coordinate choice beyond the subspace and downstream
continuation and whether the remaining interaction matters for explanation and
steering.\par}

\begin{figure}[t]
  \centering
  \IfFileExists{figures/figure1.pdf}{%
    \includegraphics[width=\textwidth]{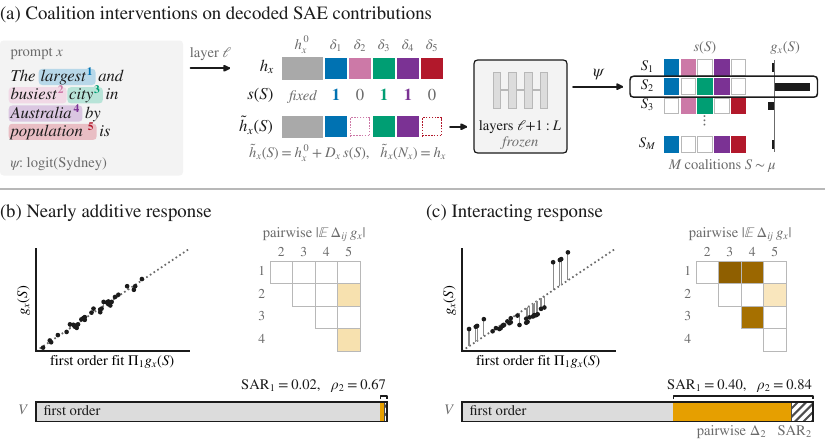}%
  }{%
    \fbox{\parbox[c][4.1cm][c]{0.94\textwidth}{\centering
    Placeholder for the SAR concept figure.}}%
  }
  \caption{Behavioral compositionality in SAE coordinates. \textbf{(a)} Selected
  decoded contributions are retained or removed while all other activation
  content is fixed, and the patched activation is propagated through the frozen
  upper layers to obtain $g_x(S)$. \textbf{(b, c)} Nearly additive and interacting
  responses, showing the first order fit, pairwise interactions, and variance
  decomposition. $\SAR_1$ is the residual beyond individual features and
  $\rho_2=\Delta_2/\SAR_1$ its pairwise share.}
  \label{fig:concept}
\end{figure}

We quantify this structure with SAE Additivity Residuals (SAR). At first order,
$\SAR_1$ is the behavioral variation not explained by individual feature
effects. Adding pairs gives $\SAR_2$, so
$\Delta_2=\SAR_1-\SAR_2$ is the structure recovered by pairs and
$\rho_2=\Delta_2/\SAR_1$ its share of the first order residual. Under
independent subset sampling, SAR has an exact interaction decomposition so that
residual structure above a given order is exactly the normalized energy of
higher order interactions. The same quantity also controls errors in preserving
the ordering of feature coalitions. Figure~\ref{fig:concept} illustrates the
intervention and contrasts nearly additive and interacting responses.

The main experiments span three model and SAE pipelines and six behavioral readouts, with additional analyses across seven pipelines and multiple representation depths. We evaluate matched coordinate rotations, first order and pairwise coalition models, combined SAE steering directions, and reconstruction quality. Our contributions are:
\begin{itemize}
\item We formulate behavioral compositionality for SAE representations and
connect low order approximation error to explicit feature interactions and
coalition ordering.

\item Using sum preserving rotations of the same selected activation subspace,
we show that compositionality under the subset intervention depends on coordinate
choice, with learned SAE coordinates supporting a lower order behavioral
description than the tested rotations.

\item We show that pairwise terms improve held out coalition ordering in each
of the 18 tested settings and outperform parameter matched random third order
controls.

\item We extend the analysis to a combined SAE steering experiment, where
pairwise modeling raises rank fidelity from 0.5874 to 0.7790, and find that
SAR and reconstruction quality differ in a matched six SAE comparison.

\end{itemize}

\section{Related Work}
\label{sec:related_work}

SAE evaluation commonly measures reconstruction, sparsity, interpretability,
and downstream utility \citep{karvonen2025saebench}. Reconstruction residuals
retain structure, concepts can be absorbed across latents, feature activity can
redistribute across prompt variants, and some model features require
multidimensional descriptions
\citep{engels2024dark,chanin2024absorption,mei2026redistribution,engels2025notall}. Polynomial SAEs
add explicit higher order decoder terms \citep{koromilas2026polysae}. These
works characterize SAE representations and reconstruction, but not whether
fixed SAE coordinates compose behaviorally after their decoded contributions
pass through the remaining model.

Work on multi-feature explanations includes semantic organization of internal
features \citep{zhang2019decision}, independent additivity
\citep{ayonrinde2024interpretability}, attribution graphs and feature circuits
\citep{ameisen2025circuit,marks2025sparse}, and interventions on internal
components
\citep{wang2023interpretability,conmy2023automated,hanna2024faith,kramar2024atp}.
Because circuits, joint explanations, and interventions combine multiple
components, individual characterization does not establish that their effects
admit a faithful low order joint description. This motivates measuring
compositionality directly in the downstream response.

Feature interactions have been studied through contextual effects in neural
networks \citep{ling2019alphago} and game-theoretic interaction indices
\citep{lundberg2017unified,sundararajan2020shapley,tsai2023faithshap,bordt2023shapley}.
SAEs provide a structured setting in which the players are learned sparse
features intended to separate activation structure into interpretable
coordinates. We study the interaction that remains when these coordinates are
combined downstream and how it is distributed across orders.
Appendix~\ref{app:indices} gives the relation to interaction indices and
coalition weighting.

\section{Behavioral Compositionality of SAE Representations}
\label{sec:method}

We first define a controlled response over subsets of selected SAE
contributions. We then measure the structure unexplained at a given interaction
order, connect it to explicit interactions and coalition ordering, and estimate
the resulting population quantities from sampled coalitions.

\subsection{From SAE features to controlled subset responses}
\label{sec:local_game}

For each prompt, we select a set of highly activated SAE features, retain or
remove different subsets of their decoded contributions, and measure how the
model's behavior changes. The remaining activation content stays fixed, so we
can study the effects of these features and their combinations.
Let $F$ be a language model and
$h_x=h_\ell(x)$ the activation at a chosen layer and token position for input
$x$. For an SAE trained at this site, let $z_i(x)$ be feature $i$'s activation
and $d_i$ its decoder direction. Let $N_x$ contain the top $K$ SAE features
by activation magnitude $|z_i(x)|$ for prompt $x$, with decoded contribution
\begin{equation}
    \delta_i(x)=z_i(x)d_i,\qquad i\in N_x.
    \label{eq:feature_contribution}
\end{equation}

For $S\subseteq N_x$, we retain the contributions in $S$ and remove the other
selected contributions,
\begin{equation}
\begin{aligned}
    \widetilde h_x(S)
    &=h_x-\sum_{i\in N_x\setminus S}\delta_i(x)
     =h_x^0+\sum_{i\in S}\delta_i(x),
\end{aligned}
\label{eq:residual_preserving_intervention}
\end{equation}
where $h_x^0=h_x-\sum_{i\in N_x}\delta_i(x)$ contains all fixed activation
content, including the reconstruction residual, so
$\widetilde h_x(N_x)=h_x$. The full coalition therefore recovers the original
activation, while the reconstruction residual and all unselected content remain
fixed, isolating behavioral changes due to the retained selected contributions.

Let $D_x$ contain the selected contributions as columns and
$s(S)\in\{0,1\}^K$ indicate those retained, so
\begin{equation}
    \widetilde h_x(S)=h_x^0+D_xs(S).
    \label{eq:sae_coordinates}
\end{equation}

For a scalar behavior $\psi$, the downstream response is
\begin{equation}
    g_x(S)=\psi\!\left(F_{\ell\leftarrow\widetilde h_x(S)}(x)\right),
    \qquad S\subseteq N_x.
    \label{eq:sae_feature_game}
\end{equation}
Here $F_{\ell\leftarrow\widetilde h_x(S)}(x)$ means replacing the activation
at the chosen layer and token position with $\widetilde h_x(S)$ and continuing
the remaining model computation. This defines a controlled set-function
response over selected SAE coordinates
while the activation remainder and downstream model remain fixed. The object of
analysis is downstream behavior rather than SAE reconstruction.

To compare learned coordinates with alternatives spanning the same selected
subspace, let $Q\in\mathbb R^{K\times K}$ be orthogonal with
$Q\mathbf 1=\mathbf 1$ and define
\begin{equation}
    D_x^{Q}=D_xQ.
    \label{eq:rotated_coordinates}
\end{equation}
Orthogonality preserves the selected subspace and $Q\mathbf 1=\mathbf 1$ its
total contribution; the fixed remainder and downstream model are unchanged.
This gives a sum preserving random coordinate baseline within the same selected
subspace.

\subsection{Measuring interaction beyond individual feature effects}
\label{sec:sar_definition}

We ask how much of the measured behavioral response can be explained by
individual SAE feature effects, and how much requires interactions. A first
order model adds individual effects; a pairwise model also captures effects
that emerge when two features are retained together. We quantify the variation
left unexplained by the best approximation of a chosen order. For fixed $x$,
write $N=N_x$ and $|N|=K$. We approximate $g_x$ using effects involving at
most $r$ selected features, with function class
\begin{equation}
    \mathcal{A}_r(N)=
    \left\{a(S)=\sum_{\substack{B\subseteq N\\|B|\le r}}
    \theta_B\ind\{B\subseteq S\}\right\}.
    \label{eq:r_additive_class}
\end{equation}
Here $\theta_B$ is the coefficient associated with feature subset $B$;
$\theta_\varnothing$ is the constant term, and $\ind\{B\subseteq S\}$
equals one when all features in $B$ are retained. These coefficients can be
fitted from the measured responses to feature combinations. Thus
$\mathcal A_1(N)$ contains the constant and individual feature terms, while
$\mathcal A_2(N)$ additionally includes pairwise terms.

Let $\mu$ denote the distribution over feature subsets and $\Pi_r g_x$ the
$L^2(\mu)$ projection of $g_x$ onto $\mathcal A_r(N)$, which minimizes
$\|g_x-a\|_\mu^2$. Define
\begin{equation}
    V(x;\psi)=\left\|g_x-\mathbb E_{S\sim\mu}g_x(S)\right\|_\mu^2.
    \label{eq:behavior_variance}
\end{equation}
The SAE Additivity Residual is
\begin{equation}
    \SAR_r(x;\psi)=
    \begin{cases}
    \dfrac{\|g_x-\Pi_r g_x\|_\mu^2}{V(x;\psi)}, & V(x;\psi)>0,\\[0.6em]
    0, & V(x;\psi)=0.
    \end{cases}
    \label{eq:sar_definition}
\end{equation}
Thus $\SAR_r$ is the fraction of behavioral variance unexplained by the best
order $r$ approximation; $\SAR_1$ is the residual beyond individual feature
effects. For example, $\SAR_1=0.05$ means that the optimal first order
approximation explains $95\%$ of the behavioral variance under $\mu$.

SAR is an approximation residual, but under independent subset sampling an
orthogonal product representation identifies this residual exactly with higher
order interaction structure. Under independent feature inclusion with
probability $p\in(0,1)$, define
\begin{equation}
    u_i(S)=\frac{\ind\{i\in S\}-p}{\sqrt{p(1-p)}},\qquad
    u_B(S)=\prod_{i\in B}u_i(S),
\end{equation}
$c_B=\mathbb E[g_x(S)u_B(S)]$, and for $S'\subseteq N\setminus B$,
\begin{equation}
    \Delta_B g_x(S')=
    \sum_{C\subseteq B}(-1)^{|B|-|C|}g_x(S'\cup C).
    \label{eq:finite_difference}
\end{equation}
For $B=\{i,j\}$, $\Delta_{ij}g_x(S')$ is the pairwise effect on background
$S'$; $c_{\{i,j\}}$ averages its signed value over backgrounds.

\begin{theorem}[Behavioral interaction decomposition]
\label{thm:sar-interactions}
If coalition memberships are independent Bernoulli variables with probability
$p\in(0,1)$, then
\begin{equation}
    c_B=[p(1-p)]^{|B|/2}\mathbb E[\Delta_B g_x(S')],
    \label{eq:coefficient-finite-difference}
\end{equation}
and, when $V(x;\psi)>0$,
\begin{equation}
    \SAR_r(x;\psi)=
    \frac{\sum_{|B|>r}[p(1-p)]^{|B|}
    \left(\mathbb E[\Delta_B g_x]\right)^2}{V(x;\psi)}.
    \label{eq:sar-interaction-energy}
\end{equation}
\end{theorem}

Under independent subset sampling, $\SAR_r$ is exactly the fraction of
behavioral variance in product measure interaction terms above order $r$;
$\SAR_1$ aggregates all such pairwise and higher order terms.

The main analysis uses the uniform distribution over all subsets, equivalent to
$p=1/2$. Then $\SAR_1$ measures interaction beyond individual coordinates and
\begin{equation}
\begin{aligned}
    \Delta_2(x;\psi)&=\SAR_1(x;\psi)-\SAR_2(x;\psi)\\
    &=\frac{1}{16V(x;\psi)}\sum_{i<j}
      \left(\mathbb E[\Delta_{ij}g_x]\right)^2,
\end{aligned}
\label{eq:delta2_interaction}
\end{equation}
with pairwise share
\begin{equation}
    \rho_2(x;\psi)=\frac{\Delta_2(x;\psi)}{\SAR_1(x;\psi)}.
    \label{eq:rho2_definition}
\end{equation}
for $\SAR_1>0$, and set $\rho_2=0$ when $\SAR_1=0$.
Here $\SAR_1$ measures interaction beyond individual effects, $\Delta_2$ the
part recovered by pairs, and $\SAR_2$ what remains beyond pairwise order. Thus
$\rho_2$ is the pairwise share in this product measure decomposition.

The projection definition also gives several basic calibration properties:
\begin{equation}
    \SAR_r=0\Longleftrightarrow g_x\in\mathcal A_r(N_x),\qquad
    0=\SAR_K\le\cdots\le\SAR_1\le1,
    \label{eq:sar_basic_properties}
\end{equation}
and, for $a\in\mathcal A_r(N_x)$,
\begin{equation}
    \frac{\|g_x-a\|_\mu^2}{V(x;\psi)}\ge\SAR_r(x;\psi).
    \label{eq:sar_lower_bound}
\end{equation}
Thus SAR decreases with interaction order, vanishes exactly when order $r$ is
sufficient, and leaves explained variance $1-\SAR_r$. Proofs are in
Appendix~\ref{app:proofs}.

First order additivity is guaranteed when the downstream continuation is affine
over the tested activation states.

\begin{corollary}[Affine continuation]
\label{cor:linear_response}
If $b_x(h)=\psi(F_{\ell\leftarrow h}(x))=c+w^\top h$ for every tested
$h=\widetilde h_x(S)$, then $\SAR_1(x;\psi)=0$.
\end{corollary}

Thus nonzero $\SAR_1$ reflects departure from affine continuation on the tested
SAE states.

\subsection{Connecting low order approximation to coalition ordering}
\label{sec:ordering}

In practice, we may want to select feature combinations that produce a
desired behavioral response. This requires knowing which combinations have
larger effects, even when their exact scores are not predicted perfectly. We
therefore measure whether a low order approximation preserves the ordering of
feature combinations. Let $S,T$ be independent coalitions and write $\Delta g=g_x(S)-g_x(T)$ and
$\Delta a=a(S)-a(T)$. Define
\begin{equation}
    \mathcal E(a)=
    \frac{\mathbb E[(\Delta g)^2\ind\{\Delta g\,\Delta a\le0\}]}{2V(x;\psi)}.
    \label{eq:weighted-ordering-error}
\end{equation}
The weighting emphasizes reversals between subsets with larger behavioral
differences and downweights near ties.

\begin{theorem}[Coalition ordering error]
\label{thm:ordering-error}
For any function $a$ on $2^N$ and $V(x;\psi)>0$,
\begin{equation}
    \mathcal E(a)\le
    \frac{\operatorname{Var}_\mu(g_x-a)}{V(x;\psi)}\le
    \frac{\|g_x-a\|_\mu^2}{V(x;\psi)}.
    \label{eq:ordering-error-bound}
\end{equation}
In particular,
\begin{equation}
    \mathcal E(\Pi_r g_x)\le\SAR_r(x;\psi).
    \label{eq:sar-ordering-bound}
\end{equation}
\end{theorem}

Thus small $\SAR_r$ guarantees small behavior weighted ordering error for the
optimal order $r$ approximation, although the converse need not hold. We examine
below its association with held out coalition ranking.

\subsection{Estimating compositionality from sampled coalitions}
\label{sec:estimation_main}

In practice, we measure the model's behavior for sampled feature combinations,
fit a low order model using some combinations, and test its predictions on
separate combinations. This avoids evaluating all $2^K$ possible subsets. We
sample from $\mu$, split into fitting and evaluation sets, and fit
\begin{equation}
    \widehat a_r=
    \argmin_{a\in\mathcal A_r(N_x)}
    \frac{1}{M_{\mathrm{fit}}}\sum_{m=1}^{M_{\mathrm{fit}}}
    \left(g_x(S_m)-a(S_m)\right)^2.
    \label{eq:monte_carlo_fit}
\end{equation}

On the evaluation set, let $\widehat L_r$ be the mean squared error and
$\widehat V$ the empirical variance of $g_x$. We report
\begin{equation}
    \widehat{\SAR}_r=\frac{\widehat L_r}{\widehat V+\epsilon}.
    \label{eq:sar-estimator-main}
\end{equation}
The fitted residual and population quantity are related by
\begin{equation}
    \|g_x-\widehat a_r\|_\mu^2=
    \|g_x-\Pi_r g_x\|_\mu^2+
    \|\widehat a_r-\Pi_r g_x\|_\mu^2.
    \label{eq:structural-estimation-decomposition}
\end{equation}
The two terms are structural residual and finite sample fitting error, so
$\widehat{\SAR}_r$ is a held out predictive residual rather than population
SAR itself.

For the evaluation set $\mathcal U$, we also report Spearman rank fidelity,
\begin{equation}
F_{\mathrm{rank}}^{(r)}=
\operatorname{Spearman}_{S\in\mathcal U}
\left(\widehat a_r(S),g_x(S)\right),
\label{eq:rank-fidelity}
\end{equation}
and denote the held out analogue of
Equation~\eqref{eq:weighted-ordering-error} by
$\widehat{\mathcal E}_r$. We additionally report sign error and selection regret.
With $e(S)=g_x(S)-g_x(N_x)$ and
$\widehat e_r(S)=\widehat a_r(S)-\widehat a_r(N_x)$, sign error is the fraction
of held out coalitions with $e(S)\widehat e_r(S)<0$. If
$S^\star=\arg\max_{S\in\mathcal U} g_x(S)$ and
$\widehat S_r=\arg\max_{S\in\mathcal U}\widehat a_r(S)$, selection regret is

\begin{equation}
R^{(r)}=
\frac{g_x(S^\star)-g_x(\widehat S_r)}
{\max_{S\in\mathcal U} g_x(S)-\min_{S\in\mathcal U} g_x(S)+\epsilon}.
\label{eq:selection-regret}
\end{equation}

All predictive metrics use coalitions unseen during fitting.
Appendix~\ref{app:estimation} reports conditioning, split and sampling stability,
and fitting budget diagnostics.

\section{Experiments}
\label{sec:experiments}

{\looseness=-1
We first characterize compositionality across behaviors and representations,
then compare learned coordinates with matched random rotations and test whether
pairwise structure recovers failures of first order explanations. We finally
examine depth, combined steering directions, and the relation to reconstruction
quality. The main comparisons use Gemma-3 4B,
Mistral-7B, and Llama-3.1 8B across simple copying, subject verb agreement,
gendered pronoun preference, CounterFact \citep{meng2022locating}, IOI
\citep{wang2023interpretability}, and greater than comparison.
The main fixed layers are 22 for Gemma-3 4B, 24 for Mistral-7B, and 21
for Llama-3.1 8B. The fixed layer landscape, rotation, fidelity, and pairwise
analyses use 1,000, 100, 800, and 150 prompts per model and behavior setting,
respectively. For each prompt, we select the top $K=16$ active SAE
contributions and sample 384 coalitions, using 256 for fitting and 128 for
evaluation. Appendix~\ref{app:experimental-specifications} gives the full
specifications; Tables~\ref{tab:pipeline-specifications},
\ref{tab:coordinate-rotation-full}, and \ref{tab:fixed-layer-1000}
report the model coverage, rotation comparisons, and fixed layer results.\par}

\subsection{Behavioral compositionality varies across behaviors and representations}
\label{sec:nonadditivity-across-pipelines}

\begin{figure}[t]
  \centering
  \includegraphics[width=0.9\textwidth]{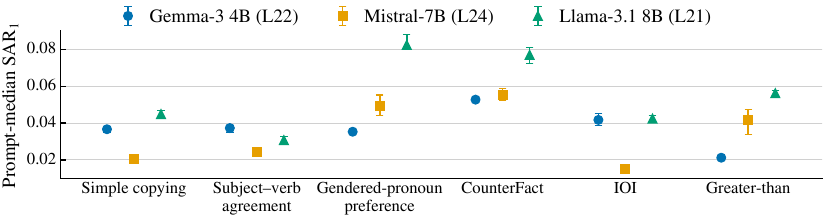}
  \caption{Prompt median $\widehat{\mathrm{SAR}}_1$ for six behaviors in
  Gemma-3 4B layer 22, Mistral-7B layer 24, and Llama-3.1 8B layer 21.
  Error bars show 95\% prompt bootstrap intervals.}
  \label{fig:behavior-landscape-1000}
\end{figure}

We first ask how well individual feature effects explain behavior across
different tasks and representations. The 57 model, layer, and behavior settings
comprise 42 settings in the three main pipelines and 15 additional Simple
Copying settings across four auxiliary pipelines. The prompt median
$\smash[t]{\widehat{\SAR}}_1$ is positive in all 57 settings while affine
controls remain at numerical zero. Many residuals are small, meaning that a
fitted first order model explains most of the measured behavioral variation.
The residual magnitude provides a direct measure of compositionality across
behaviors and representations.
Simple copying and subject verb agreement have lower residuals than CounterFact
in all three main pipelines, while the other behaviors change order across
models. The estimates therefore quantify how the degree of first order
compositionality varies with behavior and representation. The corresponding
fixed layer results are in Figure~\ref{fig:behavior-landscape-1000} and
Table~\ref{tab:fixed-layer-1000}.

\subsection{Compositionality depends on coordinate choice under subset interventions}
\label{sec:coordinate-comparison}

\begin{figure}[t]
  \centering
  \includegraphics[width=0.9\textwidth]{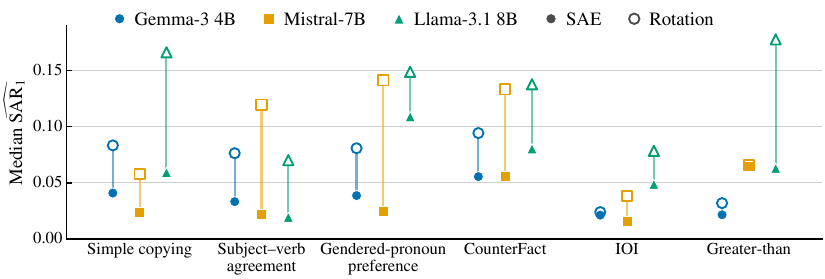}
  \caption{Paired first order SAR for learned SAE coordinates and sum preserving
  random rotations of the same selected activation subspace.}
  \label{fig:coordinate-rotation}
\end{figure}

{\looseness=-1
We compare each SAE basis with sum preserving rotations of the same selected
subspace to test whether compositionality under the subset intervention depends
on coordinate choice rather than only on the subspace and downstream
continuation.
Figure~\ref{fig:coordinate-rotation} shows the paired comparison for all 18
model and behavior combinations. Learned SAE coordinates have lower median
$\smash[t]{\widehat{\SAR}}_1$ in every comparison, although the size of the gap
varies across settings. For example, the medians are $0.0218$ and $0.1196$ for
Mistral subject verb agreement. The selected subspace, total contribution, and
downstream continuation remain fixed as the intervention coordinates change. The
systematic SAR gap therefore shows that, under this intervention, these shared
factors alone do not determine compositionality; the learned SAE basis supports
a lower order description than the tested rotations. Under milder interventions, pairwise rank improvement persists
at $p=0.75$, while halving removed contributions lowers
$\smash[t]{\widehat{\SAR}}_1$ and improves first order rank fidelity in eight of
nine settings. Appendix~\ref{app:coordinate-controls} gives the full controls.\par}

\subsection{Estimated SAR is associated with first order coalition fidelity}
\label{sec:additive-fidelity}

\begin{figure}[t]
  \centering
  \includegraphics[width=0.9\textwidth]{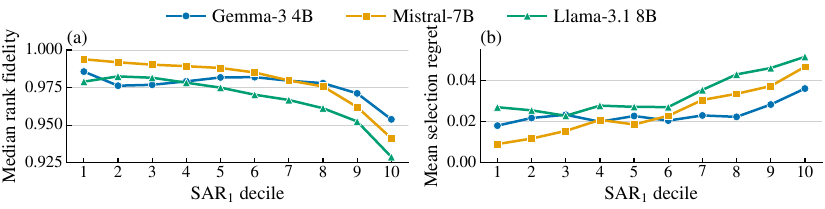}
  \caption{First order explanation fidelity across
  $\widehat{\mathrm{SAR}}_1$ deciles within each model.}
  \label{fig:additive-fidelity}
\end{figure}

{\looseness=-1
The ordering theorem bounds the population behavior weighted ordering error.
As a descriptive calibration, we compare estimated SAR with rank fidelity and
selection regret on the same held out coalitions. Figure~\ref{fig:additive-fidelity} shows lower
rank fidelity and higher selection regret from the lowest to the highest
$\smash[t]{\widehat{\SAR}}_1$ decile in all three models. Selection regret approximately
doubles in Gemma and Llama and increases by about fivefold in Mistral. Across 18
model and behavior combinations, the prompt level correlation between
$\smash[t]{\widehat{\SAR}}_1$ and rank fidelity is negative in 17 and at most $-0.6$ in
14.\par}

\subsection{Feature pairs recover missing coalition structure}
\label{sec:pairwise-improvement}

\begin{figure}[t]
  \centering
  \includegraphics[width=0.9\textwidth]{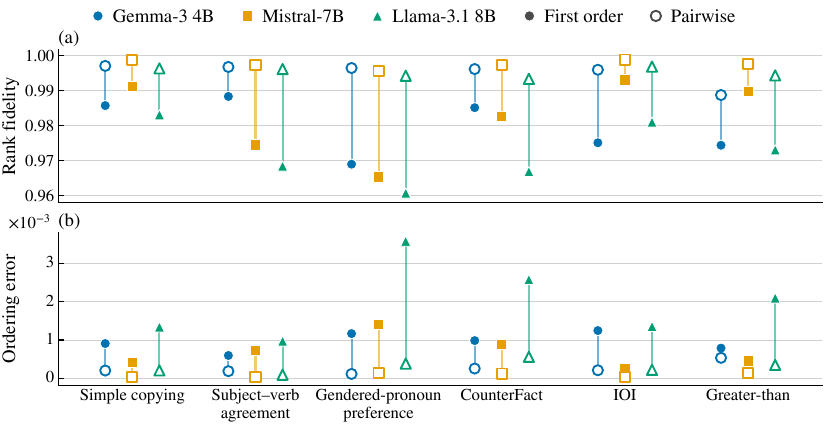}
  \caption{Held out coalition ordering for first order and pairwise
  approximations.}
  \label{fig:pairwise-rescue}
\end{figure}

{\looseness=-1
The previous results identify settings where individual feature effects are
insufficient. We therefore ask how much of the missing structure can be
recovered by the next interaction order, namely feature pairs. Figure
\ref{fig:pairwise-rescue} shows the paired comparison between first order and
pairwise approximations across all 18 model and behavior combinations.
Pairwise models improve rank fidelity in all 18 model and behavior combinations,
raising the average from 0.9782 to 0.9959, and satisfy
$\widehat{\mathcal E}_2<\widehat{\mathcal E}_1$ in all 18. Both improvements
persist across all 20 repeated fitting and evaluation splits. Selection regret
falls from 0.0272 to 0.0051 on average, and sign error also decreases in every
combination. These results show that pairwise terms capture substantial predictive
structure beyond individual feature effects. Full results are in
Appendix~\ref{app:practical-results}.\par}

{\looseness=-1
A parameter matched control retains the first order terms and replaces the
pairwise terms with the same number of randomly selected third order terms.
Pairwise models have higher held out rank fidelity and lower behavior weighted
ordering error than the median matched model in all 18 combinations. The full
comparison is in
Appendix~\ref{app:capacity-matched}.\par}

\subsection{Estimated compositionality varies across depth}
\label{sec:depth-reorganization}

\begin{figure}[t]
  \centering
  \includegraphics[width=0.9\textwidth]{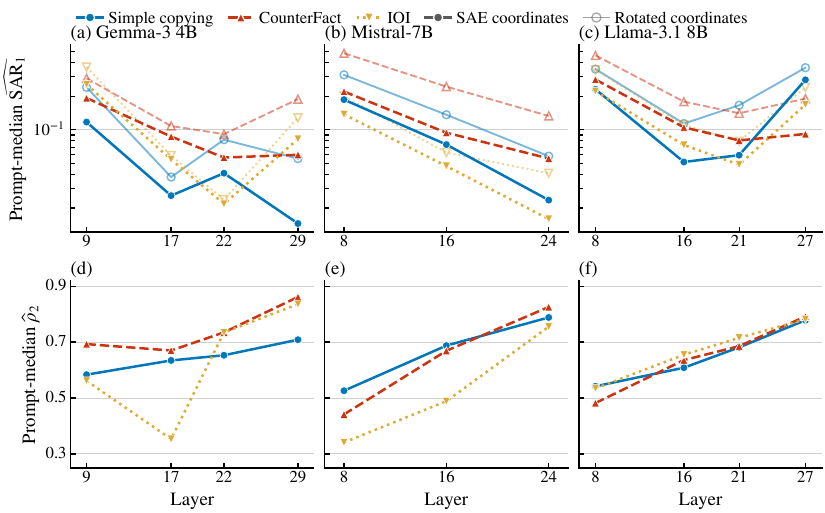}
  \caption{Depth profiles for simple copying, CounterFact, and IOI.}
  \label{fig:depth-reorganization-1000}
\end{figure}

{\looseness=-1
Layerwise analysis provides insight into how information is processed across a
network \citep{shwartz2017opening,ma2022discarding}. We next examine how the
estimated compositionality measures change across representation depth.
Figure~\ref{fig:depth-reorganization-1000} follows simple copying, CounterFact,
and IOI. Across nine early to late comparisons, $\smash[t]{\widehat{\SAR}}_1$ decreases
in eight and $\widehat{\rho}_2$ increases in all nine. Later representations therefore usually have lower estimated first order
residuals, while $\widehat{\rho}_2$ increases. This may reflect either the SAE coordinates or the shorter
remaining downstream computation. Random rotations show the same first order decrease in eight of nine
comparisons, indicating that part of the depth trend comes from the remaining
downstream computation rather than the SAE basis. SAE coordinates nevertheless
have lower $\smash[t]{\widehat{\SAR}}_1$ than rotations in all 33 model, behavior, and
layer comparisons. Thus part of the depth trend is shared across coordinate systems, while SAE
coordinates remain lower residual than rotations at matched sites;
$\widehat{\rho}_2$ also increases. The Llama
profiles are nonmonotonic near the final sampled layer.\par}

Across all seven pipelines, simple copying has lower $\smash[t]{\widehat{\SAR}}_1$ at the
latest sampled representation in six cases and higher $\widehat{\rho}_2$ in all
seven. The pattern therefore extends beyond the three main pipelines.
Appendix Figure~\ref{fig:seven-pipeline-1000} and
Table~\ref{tab:seven-pipeline-1000} give the full comparison.

\subsection{Pairwise modeling better predicts the aggregate steering response}
\label{sec:steering-ranking}

{\looseness=-1
The preceding experiments study interactions among feature contributions
already present in a model activation. SAE features are also used by adding
directions to steer model behavior. If these effects compose poorly, effects
measured for individual steering directions may not predict the outcome of
combining them. We therefore apply the same analysis to fixed published
steering directions. Using 16 published CorrSteer features \citep{cho2026corrsteer} in Llama-3.1 8B, we evaluate 1,024
steering combinations with separate fitting and evaluation coalitions. On the
overall 1,400 question MMLU-Pro response \citep{wang2024mmlupro}, rank fidelity rises from 0.5874 to 0.7790,
behavior weighted ordering error falls from 0.1444 to 0.0606, and the held out
residual estimate decreases from $0.6424$ to $0.3891$. Pairwise modeling
therefore better predicts this aggregate steering response; domain level effects are
heterogeneous and reported in Appendix~\ref{app:practical-results}.\par}

\begin{table}[!htbp]
\caption{Held out residual estimates and coalition ordering for
MMLU-Pro steering response.}
\label{tab:steering-ranking}
\centering
\small
\setlength{\tabcolsep}{7pt}
\begin{tabular}{lccc}
\toprule
Approximation & Residual estimate & Rank fidelity & Ordering error \\
\midrule
First order & 0.6424 & 0.5874 & 0.1444 \\
Pairwise & 0.3891 & 0.7790 & 0.0606 \\
\bottomrule
\end{tabular}
\end{table}

\subsection{Reconstruction quality and SAR differ in a matched SAE comparison}
\label{sec:reconstruction-comparison}

{\looseness=-1
Reconstruction quality measures how accurately an SAE reconstructs model
activations, whereas SAR measures how well feature effects explain downstream
behavior when combined. We next ask whether these properties track one another
across matched SAEs at a fixed site. Pooling prompts across six SAEs at a fixed
Gemma-3 1B site, SAR is
positively related to sign error and selection regret in all three behaviors,
while $\mathrm{FVU}_x$ is negative or near zero for sign error and near zero for
selection regret. Thus reconstruction quality and SAR capture different
variation in joint behavioral fidelity in this matched comparison. Appendix
Figure~\ref{fig:reconstruction-compositionality} gives the SAE comparison;
Appendix~\ref{app:robustness} reports sensitivity to $K$.\par}

\begin{table}[!htbp]
\caption{Prompt level Spearman correlations pooled over six SAEs
($n=1{,}800$ per behavior).}
\label{tab:reconstruction-vs-sar}
\centering
\small
\setlength{\tabcolsep}{5pt}
\begin{tabular*}{\textwidth}{@{\extracolsep{\fill}}lcccc@{}}
\toprule
& \multicolumn{2}{c}{Sign error}
& \multicolumn{2}{c}{Selection regret} \\
\cmidrule(lr){2-3}\cmidrule(lr){4-5}
Behavior & $\mathrm{FVU}_x$ & $\smash[t]{\widehat{\SAR}}_1$ & $\mathrm{FVU}_x$ & $\smash[t]{\widehat{\SAR}}_1$ \\
\midrule
Simple copying & -0.080 & 0.269 & 0.011 & 0.220 \\
Subject verb agreement & -0.114 & 0.430 & 0.005 & 0.235 \\
IOI & -0.013 & 0.396 & 0.028 & 0.199 \\
\bottomrule
\end{tabular*}
\end{table}

\section{Discussion and Conclusion}
\label{sec:discussion}

{\looseness=-1
Linear decoding makes SAE coordinates additive for activation reconstruction,
not for downstream effects. Reconstruction quality, individual feature
interpretability, and the interaction order needed for joint behavior are
therefore distinct representation properties, with SAR measuring the last
directly. This matters whenever SAE features are combined: meaningful
individual coordinates need not have additive joint behavioral effects. Learned SAE coordinates nevertheless have lower first order residuals than
the tested sum preserving random rotations of the same selected contribution
subspaces. Individual effects remain insufficient in many cases, while pairwise
terms improve coalition prediction. The SAE basis therefore supports a lower
order description than these rotations, while residual interactions remain
important for faithful joint explanations. The distinction also appears in
steering and differs from reconstruction quality in the matched SAE comparison.
SAR characterizes the behavioral compositionality of selected feature
contributions for a given prompt, behavioral readout, and intervention
distribution. This provides a concrete basis for evaluating how SAE features
combine in explanations and interventions.\par}

\subsection*{AI use statement}

In this work, we used generative AI tools to improve the readability of the manuscript, to check the correctness and internal consistency of mathematical derivations and proofs, and to assist with experimental plotting.

\subsection*{Reproducibility statement}

Population SAR, its held out estimator, the interaction decomposition, the
coalition ordering result, and the projection estimator are defined in
Section~\ref{sec:method}. Their proofs and the finite sample least squares
analysis appear in Appendices~\ref{app:proofs} and~\ref{app:estimation}. The
experimental protocol, including behavior definitions, model and SAE coverage,
feature set size, coalition sampling, and fitting and evaluation splits, is
specified in Section~\ref{sec:experiments} and
Appendix~\ref{app:experimental-specifications}. The main experiments use
$K=16$, 384 sampled coalitions per prompt, and separate fitting and evaluation
sets. The steering experiment uses 1,024 sampled coalitions with a fixed equal
split, and the coordinate and parameter matched comparisons use paired
coalition assignments. Measurement calibration appears in
Appendix~\ref{app:measurement-calibration}, and sensitivity and protocol checks
appear in Appendix~\ref{app:robustness}.

\bibliography{custom}
\bibliographystyle{iclr2027_conference}

\clearpage
\appendix

\FloatBarrier

\section{Proofs and Mathematical Details}
\label{app:proofs}

\FloatBarrier

\subsection{Notation and equivalent bases}
\label{app:notation}

\begin{table}[!htbp]
    \centering
    \small
    \caption{Notation used throughout the paper.}
    \label{tab:notation}
    \begin{tabular}{p{0.22\textwidth}p{0.70\textwidth}}
        \toprule
        Symbol & Meaning \\
        \midrule
        $x$ & Input prompt. \\
        $h_x$ & Model activation at the selected layer and token position. \\
        $z_i(x),d_i$ & Activation and decoder direction of SAE feature $i$. \\
        $\delta_i(x)$ & Decoded contribution $z_i(x)d_i$. \\
        $N_x$ & Set of top $K$ active SAE features for prompt $x$. \\
        $S$ & Coalition of retained features, $S\subseteq N_x$. \\
        $g_x$ & SAE feature intervention game. \\
        $\mu$ & Coalition distribution. \\
        $\mathcal A_r(N_x)$ & Set functions with interaction order at most $r$. \\
        $V(x;\psi)$ & Behavioral variance across interventions. \\
        $\SAR_r(x;\psi)$ & Population residual after the order $r$ projection. \\
        $\widehat{\SAR}_r$ & Held out residual estimate from the fitted order $r$ model. \\
        $\Delta_B g_x$ & Finite difference associated with feature subset $B$. \\
        $\Delta_2,\rho_2$ & Population pairwise component and its share of $\SAR_1$. \\
        $\widehat{\Delta}_2,\widehat{\rho}_2$ & Corresponding held out estimates. \\
        $\mathcal E(a)$ & Population behavior weighted coalition ordering error. \\
        $\widehat{\mathcal E}_r$ & Ordering error on held out coalitions. \\
        \bottomrule
    \end{tabular}
\end{table}

The indicator functions $\ind\{B\subseteq S\}$ with $|B|\le r$ span
$\mathcal A_r(N)$. Under product Bernoulli sampling, the functions
\begin{equation}
    u_i(S)=\frac{\ind\{i\in S\}-p}{\sqrt{p(1-p)}},
    \qquad u_B(S)=\prod_{i\in B}u_i(S)
\end{equation}
form an orthonormal basis of $L^2(2^N,\mu_p)$. The span of
$\{u_B\mid |B|\le r\}$ is the same space $\mathcal A_r(N)$. At $p=1/2$, the two
coordinate systems are related by the invertible transformation
$\ind\{i\in S\}=(1+u_i(S))/2$. This equivalence is used in the finite sample
analysis, while the finite sample estimator is expressed in the indicator basis.

\subsection{Proof of the behavioral interaction decomposition}
\label{app:interaction-proof}

Fix $B\subseteq N$ and condition on the memberships of the features outside
$B$. Let the resulting coalition be $S'\subseteq N\setminus B$. For each
$i\in B$, a present feature contributes
\begin{equation}
    p\frac{1-p}{\sqrt{p(1-p)}}=\sqrt{p(1-p)}
\end{equation}
to the conditional expectation of $u_i$, while an absent feature contributes
\begin{equation}
    (1-p)\frac{-p}{\sqrt{p(1-p)}}=-\sqrt{p(1-p)}.
\end{equation}
Expanding over the memberships inside $B$ therefore gives
\begin{align}
&\mathbb E\!\left[g_x(S)u_B(S)\mid S'\right]\\
&\quad=[p(1-p)]^{|B|/2}
\sum_{C\subseteq B}(-1)^{|B|-|C|}g_x(S'\cup C)\\
&\quad=[p(1-p)]^{|B|/2}\Delta_Bg_x(S').
\end{align}
Averaging over $S'$ proves
Equation~\eqref{eq:coefficient-finite-difference}.

Because the $u_B$ form an orthonormal basis,
\begin{equation}
    g_x(S)=\sum_{B\subseteq N}c_Bu_B(S),
    \qquad
    \|g_x-\Pi_rg_x\|_{\mu_p}^2=\sum_{|B|>r}c_B^2.
\end{equation}
The centered variance is
\begin{equation}
    V(x;\psi)=\sum_{|B|\ge1}c_B^2.
\end{equation}
Substituting the finite difference expression for $c_B$ proves
Theorem~\ref{thm:sar-interactions} and
Equation~\eqref{eq:delta2_interaction}.

\subsection{Projection properties}
\label{app:projection-properties}

If $g_x\in\mathcal A_r$, then $\Pi_rg_x=g_x$ and $\SAR_r=0$. Conversely, if
$\SAR_r=0$, full support makes $g_x$ equal to its projection on every
coalition. The nesting
\begin{equation}
    \mathcal A_0\subseteq\mathcal A_1\subseteq\cdots\subseteq\mathcal A_K
\end{equation}
then gives the residual sequence in
Equation~\eqref{eq:sar_basic_properties}.

For any $a\in\mathcal A_r$, projection orthogonality gives
\begin{equation}
    \|g_x-a\|_\mu^2
    =
    \|g_x-\Pi_rg_x\|_\mu^2
    +
    \|\Pi_rg_x-a\|_\mu^2,
    \label{eq:projection-pythagorean}
\end{equation}
which proves Equation~\eqref{eq:sar_lower_bound}. Since constants belong to
$\mathcal A_r$, the projection residual has mean zero. Applying the same
orthogonal decomposition after centering gives
\begin{equation}
    \frac{\operatorname{Var}_\mu(\Pi_rg_x)}{V(x;\psi)}=1-\SAR_r(x;\psi).
\end{equation}

If the continuation map is affine on the tested states,
\begin{align}
    g_x(S)
    &=c+w^\top\left(h_x^0+\sum_{i\in S}\delta_i(x)\right)\\
    &=c+w^\top h_x^0+\sum_{i\in S}w^\top\delta_i(x),
\end{align}
so $g_x\in\mathcal A_1$ and Corollary~\ref{cor:linear_response} follows. SAR
is also invariant to affine rescaling of $g_x$ because both its numerator and
denominator scale by the same squared factor.

\subsection{Proof of the coalition ordering theorem}
\label{app:ordering-proof}

Let
\begin{equation}
    e=g_x-a,
    \qquad
    \Delta e=e(S)-e(T)=\Delta g-\Delta a.
\end{equation}
If $\Delta g\,\Delta a\le0$ and $\Delta g\ne0$, then
$|\Delta e|\ge|\Delta g|$. Hence
\begin{equation}
    (\Delta g)^2\ind\{\Delta g\,\Delta a\le0\}
    \le(\Delta e)^2.
\end{equation}
For independent $S,T\sim\mu$,
\begin{equation}
    \mathbb E[(\Delta e)^2]=2\operatorname{Var}_\mu(e),
    \qquad
    \mathbb E[(\Delta g)^2]=2V(x;\psi).
\end{equation}
Dividing the first inequality by the second expectation gives the first bound
in Equation~\eqref{eq:ordering-error-bound}. The second follows from
$\operatorname{Var}(e)\le\mathbb E[e^2]$. When $a=\Pi_rg_x$, the residual has
mean zero and its normalized squared norm is $\SAR_r$, proving
Equation~\eqref{eq:sar-ordering-bound}.

\section{Estimation and Finite Sample Analysis}
\label{app:estimation}

\FloatBarrier

\subsection{Projection estimator}
\label{app:projection-estimator}

For each prompt, we draw coalitions with independent feature memberships and
evaluate $g_x(S_m)$. For order $r$, we use
\begin{equation}
    \phi_r(S)=\left(\ind\{B\subseteq S\}\right)_{
    B\subseteq N_x,\ |B|\le r}
\end{equation}
and fit $\widehat a_r(S)=\widehat\theta_r^\top\phi_r(S)$ by unregularized
least squares on the fitting coalitions. On the evaluation split we compute
\begin{equation}
    \widehat L_r
    =
    \frac{1}{M_{\mathrm{eval}}}
    \sum_{m=1}^{M_{\mathrm{eval}}}
    \left(g_x(S_m)-\widehat a_r(S_m)\right)^2
\end{equation}
and the held out variance $\widehat V$. The reported estimator is
\begin{equation}
    \widehat{\SAR}_r=\frac{\widehat L_r}{\widehat V+\epsilon}.
\end{equation}
The three large model pipelines use $\epsilon=10^{-8}$ and float64
projection.

\subsection{Walsh coordinates and fitting error}
\label{app:walsh-ols}

At $p=1/2$, let $\varphi_r(S)$ contain the Walsh basis terms $u_B(S)$ with
$|B|\le r$. Write
\begin{equation}
    a_r^*(S)=\Pi_rg_x(S)=\theta_*^\top\varphi_r(S),
    \qquad
    \varepsilon_r(S)=g_x(S)-a_r^*(S).
\end{equation}
The population Gram matrix is the identity and
$\mathbb E[\varphi_r(S)\varepsilon_r(S)]=0$. For a fitting sample of size $M$,
define
\begin{equation}
    \widehat G_r=\frac1M\sum_{m=1}^M
    \varphi_r(S_m)\varphi_r(S_m)^\top.
\end{equation}
When $\widehat G_r$ is invertible, the least squares normal equations give the
exact identity
\begin{equation}
    \widehat\theta_r-\theta_*
    =
    \widehat G_r^{-1}
    \frac1M\sum_{m=1}^M\varphi_r(S_m)\varepsilon_r(S_m).
    \label{eq:walsh-ols-error}
\end{equation}
Because the population Walsh basis is orthonormal,
\begin{equation}
    \|\widehat a_r-a_r^*\|_\mu^2
    =\|\widehat\theta_r-\theta_*\|_2^2.
    \label{eq:walsh-prediction-error}
\end{equation}
Equations~\eqref{eq:walsh-ols-error} and
\eqref{eq:walsh-prediction-error} show how finite sample fitting error depends
on the empirical Gram matrix and the sampled projection residual. Combining
these identities with Equation~\eqref{eq:structural-estimation-decomposition}
gives a finite sample description of the unregularized least squares estimator
used in the experiments.

The monomial and Walsh coordinates span the same function class, so a full rank
fit gives the same predictions in either basis. The Walsh form is used here
only to expose the population orthogonality and conditioning of the sampled
problem.

\subsection{Finite sample diagnostics}
\label{app:finite-sample-diagnostics}

We evaluated the fitting protocol on the 2,700 prompts used in the explanation
fidelity experiments. The first and second order models use 17 and 137
coefficients. Every sampled design is full rank. The median condition number of
the second order Walsh Gram matrix is 41.02 and its 95th percentile is 49.88.
Predictions from the monomial and Walsh coordinates agree to numerical
precision.

\begin{table}[!htbp]
\caption{Finite sample diagnostics for the 256 fitting and 128 evaluation
coalition protocol. Split variation is the standard deviation across 20 random
splits divided by the corresponding split mean. The final two rows compare a
256 coalition fit with a fit using all 384 sampled coalitions and normalize the
mean squared prediction difference by behavioral variance.}
\label{tab:finite-sample-diagnostics}
\centering
\small
\setlength{\tabcolsep}{5pt}
\begin{tabular}{lc}
\toprule
Diagnostic & Result \\
\midrule
Maximum monomial and Walsh prediction difference & $1.67\times10^{-12}$ \\
Order 1 full rank prompts & 2,700 / 2,700 \\
Order 2 full rank prompts & 2,700 / 2,700 \\
Order 2 Gram condition number, median & 41.02 \\
Order 2 Gram condition number, 95th percentile & 49.88 \\
Relative split variation of first order residual, median & 2.29\% \\
Relative split variation of second order residual, median & 2.70\% \\
Cells with pairwise rank improvement in all 20 splits & 18 / 18 \\
Cells with pairwise $\mathcal E$ reduction in all 20 splits & 18 / 18 \\
Median difference from the 384 coalition fit, order 1 & $6.91\times10^{-4}$ \\
Median difference from the 384 coalition fit, order 2 & $1.70\times10^{-3}$ \\
\bottomrule
\end{tabular}
\end{table}

The fitting budget analysis shows that the pairwise conclusion becomes stable
before the full fitting budget even though the precise second order residual
continues to change. At 224 and 256 fitting coalitions, pairwise models improve
both rank fidelity and behavior weighted ordering error in all 18 cells. The
second order rank fidelity changes by 0.14\% from 224 to 256 fitting coalitions,
while the normalized second order residual changes by about 21\%. We therefore
use the repeated split analysis to support the stability of the comparative
conclusions and do not interpret the fitted second order residual as a
high precision estimate of its population value.

We also generated five independent sets of 384 coalitions for 50 fixed prompts
in each of nine representative model and behavior combinations. All nine
combinations preserve the original qualitative first order and pairwise
conclusions across fresh coalition draws. The median within prompt coefficient
of variation is 13.7\% for $\smash[t]{\widehat{\SAR}}_1$ and 18.2\% for
$\widehat{\SAR}_2$, compared with 0.37\% and 0.16\% for first order and pairwise
rank fidelity. The ordering conclusions are therefore considerably more stable
to coalition sampling than the absolute residual magnitudes.

\begin{table}[!htbp]
\caption{Fitting budget diagnostics with the original evaluation coalitions
held fixed. The reported values aggregate the 18 model and behavior
combinations.}
\label{tab:fitting-budget}
\centering
\small
\setlength{\tabcolsep}{4pt}
\begin{tabular*}{\textwidth}{@{\extracolsep{\fill}}cccccc@{}}
\toprule
Fit coalitions & Condition & $R^{\mathrm{var}}_2$
& $F_{\mathrm{rank}}^{(2)}$ & $\mathcal E_2$ & Rank improved \\
\midrule
160 & 580.3 & 0.03071 & 0.9783 & 0.001085 & 9 / 18 \\
192 & 134.3 & 0.01516 & 0.9885 & 0.0003920 & 17 / 18 \\
224 & 65.43 & 0.01104 & 0.9915 & 0.0002436 & 18 / 18 \\
256 & 41.02 & 0.009152 & 0.9929 & 0.0001839 & 18 / 18 \\
\bottomrule
\end{tabular*}
\end{table}

\subsection{Prompt aggregation and paired endpoints}
\label{app:aggregation}

We compute $\smash[t]{\widehat{\SAR}}_1$, $\widehat{\SAR}_2$,
$\widehat{\Delta}_2$, and $\widehat{\rho}_2$ separately for each prompt.
Dataset summaries use prompt level medians unless otherwise stated. Rank
fidelity, sign error, and selection regret in the primary pairwise comparison use
prompt means. Confidence intervals resample prompts with replacement 1,000 times. Endpoint comparisons apply each
bootstrap draw to the same prompt IDs at the early and late layers.

\section{Interaction Indices and Alternative Constructions}
\label{app:indices}

\FloatBarrier

\subsection{Interaction indices and low order fits}
\label{app:indices-fits}

Interaction indices extend feature attribution from individual players to
small coalitions. The Shapley value assigns each player a weighted average of
its marginal contributions \citep{shapley1953value}, and its extensions assign
values to pairs and larger subsets
\citep{grabisch1999axiomatic,sundararajan2020shapley,tsai2023faithshap}.
Several such quantities can be interpreted as coefficients of low order fits
\citep{lundberg2017unified,tsai2023faithshap,bordt2023shapley}. SAR instead
aggregates the residual energy that remains above a chosen order.

Under product sampling with $p=1/2$, the relation to pairwise interactions is
explicit. Theorem~\ref{thm:sar-interactions} gives
\begin{equation}
    c_{\{i,j\}}
    =\tfrac14\,
    \mathbb E_{S'\subseteq N\setminus\{i,j\}}
    [\Delta_{ij}g(S')].
\end{equation}
Thus the pairwise Fourier coefficient is a scaled context averaged second
finite difference. The quantity $\Delta_2$ aggregates the squared pairwise
coefficients rather than assigning an individual importance score to each
pair.

Low order coefficients alone do not determine the residual above that order.
Under the product measure, the basis terms with $|B|\le r$ and $|B|>r$ occupy
orthogonal subspaces. Changing the latter leaves every fitted coefficient up to
order $r$ unchanged while changing $\SAR_r$. Estimating the residual therefore
requires behavioral evaluations beyond the fitted coefficients themselves,
which is why the estimator uses a separate evaluation split.

\FloatBarrier

\subsection{Coalition weighting and variance shares}
\label{app:indices-measure}

The weighting that makes a first order fit reproduce the Shapley value places
weight proportional to
$[\binom{K}{|S|}|S|(K-|S|)]^{-1}$ on nontrivial coalitions
\citep{lundberg2017unified}. This weighting correlates feature memberships, so
the product basis used in Theorem~\ref{thm:sar-interactions} is no longer
orthogonal. The differences in the SAR sequence therefore do not have the same
variance share interpretation under that weighting. Product sampling gives
independent memberships and the orthogonal decomposition used here. The two
weightings answer different questions.

\FloatBarrier

\subsection{Alternative intervention games}
\label{app:indices-games}

A natural alternative to
Equation~\eqref{eq:residual_preserving_intervention} reconstructs the
activation from the retained subset alone,
$h(S)=\sum_{i\in S}\delta_i(x)$. This discards the reconstruction residual and
the features outside the selected set. The full coalition then evaluates the
behavior of the SAE reconstruction rather than the original model activation.
The discarded component can also carry structure and behavioral effect
\citep{engels2024dark}. Equation~\eqref{eq:residual_preserving_intervention}
keeps this component fixed, anchors the full coalition at $h_x$, and makes
comparisons across SAE bases refer to the same original activation. Fixing the
surrounding context is also used in conditional coalition games for feature
attribution \citep{ma2026ssi}.

\FloatBarrier

\subsection{Scope of the zero point}
\label{app:zero-scope}

The tested states lie in
$h_x^0+\operatorname{span}\{\delta_i(x)\mid i\in N_x\}$, whose dimension is at most
$K$. A continuation can be nonlinear away from this subspace and still induce
an additive response on the tested states. Behavioral compositionality therefore
concerns the downstream computation expressed in the selected SAE coordinates.
This is the coordinate space used when those contributions form a joint
explanation or intervention.

\section{Experimental Specifications}
\label{app:experimental-specifications}

\FloatBarrier

\Needspace{0.38\textheight}
\subsection{Behavior Specifications}
\label{app:behavior-specifications}

Table~\ref{tab:behavior-specifications} lists the prompt sources and behavioral readouts used in the main experiments.

\begin{table}[!htbp]
\caption{Behavior definitions used in the main experiments.}
\label{tab:behavior-specifications}
\centering
\small
\setlength{\tabcolsep}{4pt}
\begin{tabular}{p{0.21\textwidth}p{0.30\textwidth}p{0.39\textwidth}}
\toprule
Behavior & Prompt or source & Scalar behavior $\psi$ \\
\midrule
Affine control & Simple copying prompts & Linear readout of the intervened activation \\
Simple copying & Repeated name prompts & Target name logit difference \\
Subject verb agreement & Agreement prompts ending before the verb & Correct verb logit difference \\
Gendered pronoun preference & Biography prompts ending before the pronoun & Pronoun logit difference \\
CounterFact & Factual prompts from CounterFact \citep{meng2022locating} & Target object token logit \\
IOI & Indirect object identification prompts \citep{wang2023interpretability} & Indirect object logit difference \\
Greater than & Numerical comparison prompts & Correct answer label logit difference \\
\bottomrule
\end{tabular}
\end{table}

Greater than uses answer labels that are compatible with each model tokenizer.
Gemma-3 4B and Llama-3.1 8B use A and B labels. Mistral-7B uses True and False.
The prompt templates and selected prompt identities are fixed across the
corresponding comparisons.

\FloatBarrier

\Needspace{0.42\textheight}
\subsection{Steering Specifications}
\label{app:steering-specifications}

The steering experiment uses Llama-3.1 8B with the Llama Scope residual SAEs
used by CorrSteer \citep{cho2026corrsteer,he2024llamascope}. We take the 16 MMLU-Pro
features with the largest published
positive correlations among the reported layerwise selections and keep their
published steering coefficients fixed. Table~\ref{tab:steering-features}
lists the resulting feature set.

\begin{table}[!htbp]
\caption{Fixed SAE features used in the MMLU-Pro steering experiment.}
\label{tab:steering-features}
\centering
\small
\setlength{\tabcolsep}{5pt}
\begin{tabular}{cccc}
\toprule
Index & Layer & Feature ID & Steering coefficient \\
\midrule
1 & 26 & 22133 & 0.813 \\
2 & 19 & 6432 & 0.966 \\
3 & 22 & 11286 & 0.697 \\
4 & 18 & 2624 & 0.478 \\
5 & 20 & 28406 & 0.628 \\
6 & 17 & 26604 & 0.469 \\
7 & 24 & 7967 & 0.837 \\
8 & 23 & 21146 & 0.853 \\
9 & 25 & 16619 & 0.864 \\
10 & 21 & 15538 & 0.391 \\
11 & 15 & 9446 & 0.425 \\
12 & 16 & 6219 & 0.342 \\
13 & 14 & 28135 & 0.490 \\
14 & 30 & 19427 & 0.374 \\
15 & 12 & 28750 & 0.563 \\
16 & 28 & 23202 & 1.120 \\
\bottomrule
\end{tabular}
\end{table}

A steering coalition $S$ adds the selected direction at each corresponding
layer with its fixed coefficient. Evaluation uses direct answers without
chain of thought: each question presents options A through J, and the
predicted answer is the option with the highest next-token logit among those
labels. We sample 1,024 distinct coalitions uniformly
from the $2^{16}$ possible combinations and fix an equal fitting and
evaluation split. The same coalition masks and split are
used for every MMLU-Pro domain. We sample 100 test questions from each of the
14 official MMLU-Pro domains \citep{wang2024mmlupro} and define the value of a coalition as its accuracy within
that domain. Overall MMLU-Pro accuracy is the average over the resulting 1,400
questions. The reported confidence intervals resample prompts with replacement
and recompute the accuracy games and projections.

The finite sample pairwise fit is evaluated on coalitions that were not used
for fitting. It can therefore have a larger held out residual than the first
order fit in some domains. We report $\widehat{\rho}_2$ without clipping. A
negative value records a larger held out residual for the second order fit,
while the population quantity $\rho_2$ remains nonnegative.

\FloatBarrier

\Needspace{0.38\textheight}
\subsection{Model and SAE Pipeline Specifications}
\label{app:pipeline-specifications}

Table~\ref{tab:pipeline-specifications} summarizes the model, SAE family, and layers included in the analysis. The GPT-2 and Pythia SAEs follow the releases described by \citet{bloom2024saelens}; the Gemma-3 SAEs use Gemma Scope 2 \citep{mcdougall2025gemmascope2}, the Mistral SAEs use the residual SAEs of \citet{engels2025notall}, and the Llama SAEs use Llama Scope \citep{he2024llamascope}.

\begin{table}[!htbp]
\caption{Model and SAE pipeline coverage.}
\label{tab:pipeline-specifications}
\centering
\small
\setlength{\tabcolsep}{7pt}
\begin{tabular}{@{}llc@{}}
\toprule
Model & SAE family & Layers analyzed \\
\midrule
GPT-2 Small & Bloom GPT-2 SAE & 4, 6, 8, 10 \\
Pythia-70M & Bloom Pythia SAE & 1, 3, 5 \\
Gemma-3 270M & Gemma Scope 2 & 5, 9, 12, 15 \\
Gemma-3 1B & Gemma Scope 2 & 7, 13, 17, 22 \\
Gemma-3 4B & Gemma Scope 2 & 9, 17, 22, 29 \\
Mistral-7B & Engels residual SAE & 8, 16, 24 \\
Llama-3.1 8B & Llama Scope & 8, 16, 21, 27 \\
\bottomrule
\end{tabular}
\end{table}

For each prompt, $N_x$ contains the top $K$ active features by activation
magnitude. The released decoding convention of each SAE is used when converting
feature activations to decoder contributions. The three main pipelines use
unquantized model evaluation. Prompt identities, selected feature identities,
coalition assignments, and fitting and evaluation splits are fixed within each
comparison.

\FloatBarrier

\section{Measurement Calibration}
\label{app:measurement-calibration}

Table~\ref{tab:synthetic-checks} reports synthetic checks of the residual estimator before the affine controls in the language models.

\begin{table}[!htbp]
\caption{Synthetic checks for the SAR estimator.}
\label{tab:synthetic-checks}
\centering
\small
\begin{tabular}{lcc}
\toprule
Game & $\widehat{\mathrm{SAR}}_1$ & $\widehat{\mathrm{SAR}}_2$ \\
\midrule
Additive & $6.1\times10^{-12}$ & $2.6\times10^{-10}$ \\
Pairwise with target residual 0.1 & 0.100 & $\approx0$ \\
Pairwise with target residual 0.3 & 0.300 & $\approx0$ \\
Pairwise with target residual 0.5 & 0.500 & $\approx0$ \\
Third order & 0.180 & 0.066 \\
\bottomrule
\end{tabular}
\end{table}

Across the 11 affine controls in Gemma-3 4B, Mistral-7B, and Llama-3.1 8B,
the largest prompt median $\widehat{\mathrm{SAR}}_1$ is $1.55\times10^{-13}$ with
1,000 prompts for each control.

\FloatBarrier

\section{Coordinate Comparisons}
\label{app:coordinate-controls}

\FloatBarrier

The same selected activation subspace can be written with different coordinate
decompositions through Equation~\eqref{eq:rotated_coordinates}. For each prompt,
we sample eight rotations independently by fixing the normalized all ones
direction and QR orthogonalizing a Gaussian matrix on its $(K-1)$ dimensional
orthogonal complement. Table~\ref{tab:coordinate-rotation-full} reports the main
fixed layer comparison.

\begin{table}[!htbp]
\caption{Estimated first order SAR for learned SAE coordinates and random
coordinate rotations within the same selected activation subspace. Each row
uses the first 100 prompts in the dataset's fixed stratum ordering and eight
rotations per prompt. The final column gives the fraction of prompts whose
SAE estimate lies below the median of its rotations. These 100 prompt estimates
are distinct from the 1,000 prompt summaries in Table~\ref{tab:fixed-layer-1000}.}
\label{tab:coordinate-rotation-full}
\centering
\footnotesize
\setlength{\tabcolsep}{2.8pt}
\begin{tabular*}{\textwidth}{@{\extracolsep{\fill}}llccc@{}}
\toprule
Model & Behavior & SAE $\smash[t]{\widehat{\SAR}}_1$ & Rotation $\smash[t]{\widehat{\SAR}}_1$ & SAE below rotation median \\
\midrule
Gemma-3 4B & Simple copying & 0.04097 & 0.08333 & 0.98 \\
& Subject verb agreement & 0.03332 & 0.07638 & 0.99 \\
& Gendered pronoun preference & 0.03872 & 0.08080 & 1.00 \\
& CounterFact & 0.05558 & 0.09428 & 0.92 \\
& IOI & 0.02132 & 0.02402 & 0.66 \\
& Greater than & 0.02165 & 0.03198 & 0.98 \\
\midrule
Mistral-7B & Simple copying & 0.02348 & 0.05794 & 0.98 \\
& Subject verb agreement & 0.02179 & 0.1196 & 1.00 \\
& Gendered pronoun preference & 0.02428 & 0.1411 & 0.98 \\
& CounterFact & 0.05545 & 0.1331 & 1.00 \\
& IOI & 0.01589 & 0.03812 & 1.00 \\
& Greater than & 0.06456 & 0.06556 & 0.54 \\
\midrule
Llama-3.1 8B & Simple copying & 0.05932 & 0.1661 & 0.99 \\
& Subject verb agreement & 0.01944 & 0.07024 & 0.99 \\
& Gendered pronoun preference & 0.1089 & 0.1488 & 0.87 \\
& CounterFact & 0.08039 & 0.1377 & 0.90 \\
& IOI & 0.04875 & 0.07851 & 0.95 \\
& Greater than & 0.06296 & 0.1776 & 1.00 \\
\bottomrule
\end{tabular*}
\end{table}

\FloatBarrier

We also varied the intervention without changing the selected feature identities.
Across nine representative model and behavior combinations, increasing feature
inclusion from $p=0.5$ to $p=0.75$ preserves pairwise rank improvement in all
nine combinations. Reducing each removed contribution to half of its observed
magnitude lowers $\smash[t]{\widehat{\SAR}}_1$ in eight of nine combinations and increases
first order rank fidelity in eight of nine. At smaller perturbation magnitude,
the downstream response is therefore closer to a first order description.

\FloatBarrier

\section{Full Behavioral Compositionality Results}
\label{app:full-sar-results}

\FloatBarrier

\Needspace{0.58\textheight}
\subsection{Fixed Layer Behavior Comparisons}

Table~\ref{tab:fixed-layer-1000} gives the fixed layer estimates corresponding to the behavior comparison in the main text.

\begin{table}[!htbp]
\caption{Fixed layer behavior estimates using the full 1,000 prompts for each
combination, rather than the 100 prompt rotation subset in
Table~\ref{tab:coordinate-rotation-full}. Brackets give 95\% confidence
intervals from prompt bootstrap resampling.}
\label{tab:fixed-layer-1000}
\centering
\footnotesize
\setlength{\tabcolsep}{0.8pt}
\begin{tabular*}{\textwidth}{@{\extracolsep{\fill}}llccc@{}}
\toprule
Pipeline and layer & Behavior & $n$ & Median $\smash[t]{\widehat{\SAR}}_1$ & Median $\widehat{\rho}_2$ \\
\midrule
Gemma-3 4B L22 & Simple copying & 1000 & 0.03668 [0.03523, 0.03791] & 0.6530 [0.6368, 0.6687] \\
& Subject verb agreement & 1000 & 0.03732 [0.03496, 0.03877] & 0.6274 [0.5978, 0.6518] \\
& Gendered pronoun preference & 1000 & 0.03532 [0.03426, 0.03634] & 0.8296 [0.8247, 0.8348] \\
& CounterFact & 1000 & 0.05275 [0.05128, 0.05423] & 0.7348 [0.7190, 0.7447] \\
& IOI & 1000 & 0.04174 [0.03867, 0.04507] & 0.7346 [0.7229, 0.7412] \\
& Greater than & 1000 & 0.02121 [0.02090, 0.02162] & 0.2814 [0.2653, 0.2949] \\
\midrule
Mistral-7B L24 & Simple copying & 1000 & 0.02029 [0.01953, 0.02115] & 0.7882 [0.7783, 0.7944] \\
& Subject verb agreement & 1000 & 0.02435 [0.02304, 0.02538] & 0.8718 [0.8503, 0.8842] \\
& Gendered pronoun preference & 1000 & 0.04939 [0.04431, 0.05533] & 0.8166 [0.8070, 0.8272] \\
& CounterFact & 1000 & 0.05535 [0.05259, 0.05862] & 0.8255 [0.8193, 0.8328] \\
& IOI & 1000 & 0.01510 [0.01460, 0.01577] & 0.7552 [0.7452, 0.7628] \\
& Greater than & 1000 & 0.04159 [0.03366, 0.04756] & 0.7804 [0.7757, 0.7854] \\
\midrule
Llama-3.1 8B L21 & Simple copying & 1000 & 0.04486 [0.04313, 0.04698] & 0.6813 [0.6739, 0.6885] \\
& Subject verb agreement & 1000 & 0.03070 [0.02853, 0.03253] & 0.7837 [0.7716, 0.7937] \\
& Gendered pronoun preference & 1000 & 0.08244 [0.07991, 0.08794] & 0.7778 [0.7709, 0.7864] \\
& CounterFact & 1000 & 0.07707 [0.07241, 0.08102] & 0.6855 [0.6713, 0.6998] \\
& IOI & 1000 & 0.04245 [0.04046, 0.04401] & 0.7152 [0.7078, 0.7204] \\
& Greater than & 1000 & 0.05625 [0.05505, 0.05786] & 0.7085 [0.6931, 0.7231] \\
\bottomrule
\end{tabular*}
\end{table}

\FloatBarrier

\Needspace{0.48\textheight}
\subsection{Paired Early and Late Comparisons}

Table~\ref{tab:endpoints-1000} reports the paired endpoints used to summarize the depth profiles.

\begin{table}[!htbp]
\caption{Paired early to late estimates for the three detailed depth profiles.
Brackets in the reduction column give 95\% confidence intervals from paired
prompt bootstrap resampling.}
\label{tab:endpoints-1000}
\centering
\small
\setlength{\tabcolsep}{2.5pt}
\begin{tabular}{llccc}
\toprule
Pipeline & Behavior & Early $\to$ late $\smash[t]{\widehat{\SAR}}_1$ & Reduction & Early $\to$ late $\widehat{\rho}_2$ \\
\midrule
Gemma-3 4B & Simple copying & 0.1122 $\to$ 0.01155 & 9.711 [9.147, 10.29] & 0.5835 $\to$ 0.7085 \\
& CounterFact & 0.1723 $\to$ 0.07646 & 2.254 [2.112, 2.483] & 0.6929 $\to$ 0.8614 \\
& IOI & 0.2255 $\to$ 0.07320 & 3.080 [2.948, 3.248] & 0.5615 $\to$ 0.8358 \\
Mistral-7B & Simple copying & 0.1771 $\to$ 0.02029 & 8.728 [8.232, 9.242] & 0.5261 $\to$ 0.7882 \\
& CounterFact & 0.2658 $\to$ 0.05535 & 4.801 [4.432, 5.201] & 0.4415 $\to$ 0.8255 \\
& IOI & 0.1270 $\to$ 0.01510 & 8.412 [7.865, 8.934] & 0.3411 $\to$ 0.7552 \\
Llama-3.1 8B & Simple copying & 0.2001 $\to$ 0.2562 & 0.781 [0.737, 0.831] & 0.5419 $\to$ 0.7782 \\
& CounterFact & 0.2817 $\to$ 0.09233 & 3.051 [2.872, 3.305] & 0.4822 $\to$ 0.7916 \\
& IOI & 0.2151 $\to$ 0.1769 & 1.216 [1.123, 1.303] & 0.5345 $\to$ 0.7813 \\
\bottomrule
\end{tabular}
\end{table}

\FloatBarrier

\Needspace{0.58\textheight}
\subsection{Simple Copying Across Seven Pipelines}

\begin{figure}[!htbp]
  \centering
  \includegraphics[width=0.98\textwidth]{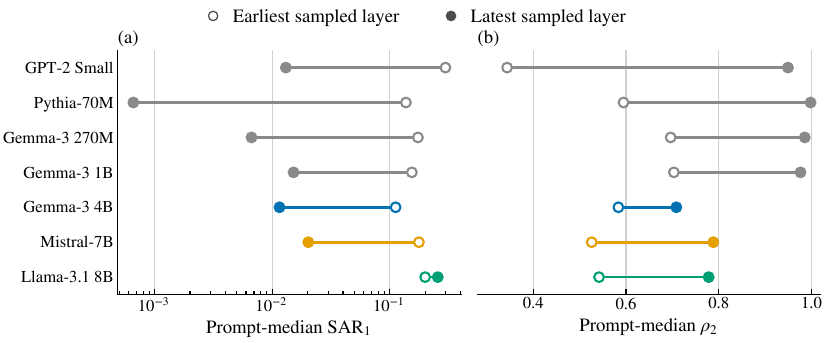}
  \caption{Simple copying at the earliest and latest sampled layers across
  seven model and SAE pipelines. Open markers denote the earliest layer and
  filled markers denote the latest layer. The left panel reports prompt median
  $\widehat{\mathrm{SAR}}_1$ and the right panel reports prompt median
  $\widehat{\rho}_2$.}
  \label{fig:seven-pipeline-1000}
\end{figure}

Figure~\ref{fig:seven-pipeline-1000} and
Table~\ref{tab:seven-pipeline-1000} give the full comparison across seven
pipelines referenced in the main text.

\begin{table}[!htbp]
\caption{Estimated simple copying endpoints across seven model and SAE pipelines.}
\label{tab:seven-pipeline-1000}
\centering
\small
\setlength{\tabcolsep}{4pt}
\begin{tabular*}{\textwidth}{@{\extracolsep{\fill}}lcccc@{}}
\toprule
Pipeline & Early layer & Late layer & Early $\to$ late $\smash[t]{\widehat{\SAR}}_1$ & Early $\to$ late $\widehat{\rho}_2$ \\
\midrule
GPT-2 Small & 4 & 10 & 0.2978 $\to$ 0.01309 & 0.3437 $\to$ 0.9491 \\
Pythia-70M & 1 & 5 & 0.1373 $\to$ 0.0006608 & 0.5946 $\to$ 0.9979 \\
Gemma-3 270M & 5 & 15 & 0.1738 $\to$ 0.006676 & 0.6962 $\to$ 0.9854 \\
Gemma-3 1B & 7 & 22 & 0.1543 $\to$ 0.01521 & 0.7032 $\to$ 0.9763 \\
Gemma-3 4B & 9 & 29 & 0.1122 $\to$ 0.01155 & 0.5835 $\to$ 0.7085 \\
Mistral-7B & 8 & 24 & 0.1771 $\to$ 0.02029 & 0.5261 $\to$ 0.7882 \\
Llama-3.1 8B & 8 & 27 & 0.2001 $\to$ 0.2562 & 0.5419 $\to$ 0.7782 \\
\bottomrule
\end{tabular*}
\end{table}

\FloatBarrier

\Needspace{0.42\textheight}
\subsection{Greater Than Depth Profiles}

Table~\ref{tab:gt-depth} reports the complete greater than depth measurements available for Gemma and Llama.

\begin{table}[!htbp]
\caption{Estimated greater than depth profiles in the two pipelines with complete depth measurements.}
\label{tab:gt-depth}
\centering
\small
\begin{tabular}{lcccc}
\toprule
Pipeline & Layer & Median $\smash[t]{\widehat{\SAR}}_1$ & Median $\widehat{\SAR}_2$ & Median $\widehat{\rho}_2$ \\
\midrule
Gemma-3 4B & 9 & 0.2045 & 0.07585 & 0.6223 \\
& 17 & 0.1436 & 0.09039 & 0.3809 \\
& 22 & 0.02121 & 0.01495 & 0.2814 \\
& 29 & 0.06386 & 0.009314 & 0.8551 \\
Llama-3.1 8B & 8 & 0.3227 & 0.1005 & 0.6623 \\
& 16 & 0.1302 & 0.03133 & 0.7385 \\
& 21 & 0.05625 & 0.01509 & 0.7085 \\
& 27 & 0.2147 & 0.04577 & 0.7822 \\
\bottomrule
\end{tabular}
\end{table}

\FloatBarrier

\section{Explanation Fidelity Results}
\label{app:practical-results}

For a held out coalition $S$, let
\[
    e(S)=g_x(S)-g_x(N_x),
    \qquad
    \widehat e_r(S)=\widehat a_r(S)-\widehat a_r(N_x).
\]
Sign error is the fraction of evaluated coalitions for which $e(S)$ and
$\widehat e_r(S)$ have opposite signs. For the same evaluation set $\mathcal U$, let
$S^\star=\arg\max_{S\in\mathcal U} g_x(S)$ and
$\widehat S_r=\arg\max_{S\in\mathcal U}\widehat a_r(S)$. Selection regret,
introduced in Equation~\eqref{eq:selection-regret}, is
\begin{equation}
    R^{(r)}
    =
    \frac{g_x(S^\star)-g_x(\widehat S_r)}
    {\max_{S\in\mathcal U} g_x(S)-\min_{S\in\mathcal U} g_x(S)+\epsilon}.
\end{equation}
These measures supplement rank fidelity and behavior weighted ordering error.

\FloatBarrier

\Needspace{0.42\textheight}
\subsection{Estimated SAR and First Order Fidelity}

Table~\ref{tab:decile-endpoints} gives the endpoint values for the
$\smash[t]{\widehat{\SAR}}_1$ decile comparison in Figure~\ref{fig:additive-fidelity}.

\begin{table}[!htbp]
\caption{First order explanation fidelity in the lowest and highest
$\smash[t]{\widehat{\SAR}}_1$ deciles within each model.}
\label{tab:decile-endpoints}
\centering
\small
\setlength{\tabcolsep}{4pt}
\begin{tabular*}{\textwidth}{@{\extracolsep{\fill}}lcccc@{}}
\toprule
& \multicolumn{2}{c}{Rank fidelity} & \multicolumn{2}{c}{Selection regret} \\
\cmidrule(lr){2-3}\cmidrule(lr){4-5}
Model & Lowest decile & Highest decile & Lowest decile & Highest decile \\
\midrule
Gemma-3 4B & 0.9857 & 0.9539 & 0.01812 & 0.03623 \\
Mistral-7B & 0.9939 & 0.9413 & 0.009118 & 0.04684 \\
Llama-3.1 8B & 0.9790 & 0.9289 & 0.02711 & 0.05173 \\
\bottomrule
\end{tabular*}
\end{table}

\FloatBarrier

\Needspace{0.60\textheight}
\subsection{Pairwise Approximation}

Tables~\ref{tab:pairwise-full} and~\ref{tab:pairwise-rank-full} report the
complete first order and pairwise results for the 18 model and behavior
combinations.

\begin{table}[!htbp]
\caption{First order and pairwise explanation errors for all 18 model and behavior combinations.}
\label{tab:pairwise-full}
\centering
\small
\setlength{\tabcolsep}{2.2pt}
\begin{tabular*}{\textwidth}{@{\extracolsep{\fill}}llcccc@{}}
\toprule
& & \multicolumn{2}{c}{Sign error} & \multicolumn{2}{c}{Selection regret} \\
\cmidrule(lr){3-4}\cmidrule(lr){5-6}
Model & Behavior & First order & Pairwise & First order & Pairwise \\
\midrule
Gemma-3 4B & Simple copying & 0.0892 & 0.0420 & 0.0150 & 0.0044 \\
& Subject verb agreement & 0.0784 & 0.0566 & 0.0175 & 0.0083 \\
& Gendered pronoun preference & 0.1070 & 0.0415 & 0.0261 & 0.0050 \\
& CounterFact & 0.0457 & 0.0418 & 0.0176 & 0.0034 \\
& IOI & 0.1275 & 0.0514 & 0.0288 & 0.0041 \\
& Greater than & 0.1288 & 0.1072 & 0.0269 & 0.0090 \\
\midrule
Mistral-7B & Simple copying & 0.0451 & 0.0231 & 0.0185 & 0.0016 \\
& Subject verb agreement & 0.1199 & 0.0482 & 0.0328 & 0.0032 \\
& Gendered pronoun preference & 0.0864 & 0.0487 & 0.0301 & 0.0023 \\
& CounterFact & 0.0960 & 0.0684 & 0.0251 & 0.0037 \\
& IOI & 0.0474 & 0.0259 & 0.0101 & 0.0018 \\
& Greater than & 0.1231 & 0.0608 & 0.0350 & 0.0044 \\
\midrule
Llama-3.1 8B & Simple copying & 0.0730 & 0.0433 & 0.0275 & 0.0052 \\
& Subject verb agreement & 0.0957 & 0.0500 & 0.0487 & 0.0077 \\
& Gendered pronoun preference & 0.0997 & 0.0612 & 0.0370 & 0.0109 \\
& CounterFact & 0.0861 & 0.0564 & 0.0378 & 0.0060 \\
& IOI & 0.0748 & 0.0423 & 0.0195 & 0.0036 \\
& Greater than & 0.0985 & 0.0484 & 0.0361 & 0.0065 \\
\bottomrule
\end{tabular*}
\end{table}

\FloatBarrier

Table~\ref{tab:pairwise-rank-full} reports the corresponding rank fidelity
values for the same model and behavior combinations.

\begin{table}[!htbp]
\caption{Mean prompt level rank fidelity of the first order and pairwise
approximations for the 18 model and behavior combinations.}
\label{tab:pairwise-rank-full}
\centering
\small
\setlength{\tabcolsep}{5pt}
\begin{tabular*}{\textwidth}{@{\extracolsep{\fill}}llcc@{}}
\toprule
Model & Behavior & $F_{\mathrm{rank}}^{(1)}$ & $F_{\mathrm{rank}}^{(2)}$ \\
\midrule
Gemma-3 4B & Simple copying & 0.9857 & 0.9970 \\
& Subject verb agreement & 0.9883 & 0.9967 \\
& Gendered pronoun preference & 0.9690 & 0.9964 \\
& CounterFact & 0.9851 & 0.9961 \\
& IOI & 0.9751 & 0.9959 \\
& Greater than & 0.9744 & 0.9887 \\
\midrule
Mistral-7B & Simple copying & 0.9912 & 0.9987 \\
& Subject verb agreement & 0.9745 & 0.9972 \\
& Gendered pronoun preference & 0.9654 & 0.9956 \\
& CounterFact & 0.9826 & 0.9973 \\
& IOI & 0.9930 & 0.9988 \\
& Greater than & 0.9897 & 0.9975 \\
\midrule
Llama-3.1 8B & Simple copying & 0.9831 & 0.9963 \\
& Subject verb agreement & 0.9685 & 0.9961 \\
& Gendered pronoun preference & 0.9608 & 0.9942 \\
& CounterFact & 0.9669 & 0.9933 \\
& IOI & 0.9810 & 0.9968 \\
& Greater than & 0.9731 & 0.9943 \\
\bottomrule
\end{tabular*}
\end{table}

\FloatBarrier

\Needspace{0.55\textheight}
\subsection{Behavior Weighted Ordering Error}

Table~\ref{tab:ordering-error-full} reports the behavior weighted ordering
errors used in the main text. Values use the original fitting and evaluation
split.

\begin{table}[!htbp]
\caption{Held out behavior weighted ordering error for first order and pairwise
approximations in the 18 model and behavior combinations used in the fidelity
analysis.}
\label{tab:ordering-error-full}
\centering
\small
\setlength{\tabcolsep}{4pt}
\begin{tabular*}{\textwidth}{@{\extracolsep{\fill}}llcc@{}}
\toprule
Model & Behavior & $\widehat{\mathcal E}_1$ & $\widehat{\mathcal E}_2$ \\
\midrule
Gemma-3 4B & Simple copying & 0.0009103 & 0.0002062 \\
& Subject verb agreement & 0.0005988 & 0.0001896 \\
& Gendered pronoun preference & 0.001169 & 0.0001176 \\
& CounterFact & 0.0009869 & 0.0002584 \\
& IOI & 0.001248 & 0.0002133 \\
& Greater than & 0.0007911 & 0.0005375 \\
\midrule
Mistral-7B & Simple copying & 0.0004209 & 0.00004362 \\
& Subject verb agreement & 0.0007353 & 0.00003234 \\
& Gendered pronoun preference & 0.001413 & 0.0001367 \\
& CounterFact & 0.0008927 & 0.0001262 \\
& IOI & 0.0002645 & 0.00003632 \\
& Greater than & 0.0004570 & 0.0001393 \\
\midrule
Llama-3.1 8B & Simple copying & 0.001344 & 0.0002101 \\
& Subject verb agreement & 0.0009746 & 0.00009669 \\
& Gendered pronoun preference & 0.003579 & 0.0003866 \\
& CounterFact & 0.002580 & 0.0005645 \\
& IOI & 0.001357 & 0.0002247 \\
& Greater than & 0.002097 & 0.0003523 \\
\bottomrule
\end{tabular*}
\end{table}

\FloatBarrier

\Needspace{0.58\textheight}
\subsection{Pairwise Terms and Matched Higher Order Capacity}
\label{app:capacity-matched}

The pairwise model contains 137 coefficients. For the matched comparison, each
control model keeps the intercept and 16 first order Walsh terms and replaces
the 120 pairwise terms with 120 terms sampled uniformly without replacement
from the 560 third order Walsh terms. Fifty control models are fitted for every
prompt with the same fitting and evaluation coalitions as the pairwise model.

\begin{table}[!htbp]
\caption{Held out ordering for the pairwise model and parameter matched models
that replace pairwise terms with randomly sampled third order Walsh terms. All
columns report prompt medians; matched quantities first take the median over 50
random models for each prompt.}
\label{tab:capacity-matched}
\centering
\footnotesize
\setlength{\tabcolsep}{2.4pt}
\begin{tabular*}{\textwidth}{@{\extracolsep{\fill}}llcccc@{}}
\toprule
Model & Behavior & Pairwise rank & Matched rank & Pairwise $\widehat{\mathcal E}$ & Matched $\widehat{\mathcal E}$ \\
\midrule
Gemma-3 4B & Simple copying & 0.9934 & 0.9802 & 0.0002062 & 0.001488 \\
& Subject verb agreement & 0.9934 & 0.9845 & 0.0001896 & 0.0009812 \\
& Gendered pronoun preference & 0.9925 & 0.9597 & 0.0001176 & 0.001838 \\
& CounterFact & 0.9926 & 0.9772 & 0.0002584 & 0.001730 \\
& IOI & 0.9925 & 0.9690 & 0.0002133 & 0.001799 \\
& Greater than & 0.9800 & 0.9682 & 0.0005375 & 0.001151 \\
\midrule
Mistral-7B & Simple copying & 0.9974 & 0.9885 & 0.00004362 & 0.0006113 \\
& Subject verb agreement & 0.9943 & 0.9674 & 0.00003234 & 0.001186 \\
& Gendered pronoun preference & 0.9912 & 0.9530 & 0.0001367 & 0.002338 \\
& CounterFact & 0.9946 & 0.9760 & 0.0001262 & 0.001579 \\
& IOI & 0.9976 & 0.9904 & 0.00003632 & 0.0004264 \\
& Greater than & 0.9949 & 0.9847 & 0.0001393 & 0.0007631 \\
\midrule
Llama-3.1 8B & Simple copying & 0.9933 & 0.9766 & 0.0002101 & 0.001939 \\
& Subject verb agreement & 0.9924 & 0.9593 & 0.00009669 & 0.001486 \\
& Gendered pronoun preference & 0.9899 & 0.9517 & 0.0003866 & 0.005213 \\
& CounterFact & 0.9864 & 0.9578 & 0.0005645 & 0.003916 \\
& IOI & 0.9937 & 0.9756 & 0.0002247 & 0.002034 \\
& Greater than & 0.9898 & 0.9661 & 0.0003523 & 0.003020 \\
\bottomrule
\end{tabular*}
\end{table}

\FloatBarrier

\Needspace{0.58\textheight}
\subsection{Steering Results Across MMLU-Pro Domains}

Table~\ref{tab:steering-ranking-full} gives the steering results for each domain summarized in the main text.

\begin{table}[!htbp]
\caption{Held out residual estimates and rank fidelity for steering
combinations on MMLU-Pro. The first order and pairwise models are fitted on one
set of sampled coalitions and evaluated on a separate set. The final column reports
$F_{\mathrm{rank}}^{(2)}-F_{\mathrm{rank}}^{(1)}$.}
\label{tab:steering-ranking-full}
\centering
\small
\setlength{\tabcolsep}{1.8pt}
\begin{tabular}{@{}lcccccc@{}}
\toprule
Domain & $\smash[t]{\widehat{\SAR}}_1$ & $F_{\mathrm{rank}}^{(1)}$ & $\widehat{\SAR}_2$ & $\widehat{\rho}_2$ & $F_{\mathrm{rank}}^{(2)}$ & Rank fidelity change \\
\midrule
Biology & 0.5815 & 0.6463 & 0.5600 & 0.0370 & 0.6684 & +0.0221 \\
Business & 0.6666 & 0.5417 & 0.3878 & 0.4183 & 0.6784 & +0.1367 \\
Chemistry & 0.9179 & 0.3170 & 0.5650 & 0.3844 & 0.7130 & +0.3960 \\
Computer Science & 0.5420 & 0.6841 & 0.5356 & 0.0119 & 0.6962 & +0.0121 \\
Economics & 0.5781 & 0.6558 & 0.6150 & -0.0639 & 0.6465 & -0.0093 \\
Engineering & 0.2238 & 0.8731 & 0.1871 & 0.1637 & 0.8689 & -0.0042 \\
Health & 0.8685 & 0.3698 & 0.6871 & 0.2089 & 0.5922 & +0.2224 \\
History & 0.6950 & 0.5716 & 0.4261 & 0.3869 & 0.7635 & +0.1919 \\
Law & 0.7685 & 0.5062 & 0.9275 & -0.2069 & 0.4155 & -0.0907 \\
Math & 0.6151 & 0.6244 & 0.6543 & -0.0638 & 0.5988 & -0.0256 \\
Other & 0.4958 & 0.6889 & 0.4672 & 0.0577 & 0.6976 & +0.0087 \\
Philosophy & 0.4549 & 0.7750 & 0.3637 & 0.2005 & 0.8039 & +0.0289 \\
Physics & 0.9071 & 0.3133 & 0.9421 & -0.0387 & 0.3669 & +0.0536 \\
Psychology & 0.5055 & 0.6956 & 0.5085 & -0.0060 & 0.7314 & +0.0358 \\
\midrule
Overall & 0.6424 & 0.5874 & 0.3891 & 0.3942 & 0.7790 & +0.1916 \\
\bottomrule
\end{tabular}
\end{table}

\begin{table}[!htbp]
\caption{Held out behavior weighted ordering error for the steering responses.}
\label{tab:steering-ordering-error}
\centering
\small
\setlength{\tabcolsep}{6pt}
\begin{tabular}{lcc}
\toprule
Domain & $\widehat{\mathcal E}_1$ & $\widehat{\mathcal E}_2$ \\
\midrule
Biology & 0.1187 & 0.1080 \\
Business & 0.1287 & 0.06926 \\
Chemistry & 0.3078 & 0.09323 \\
Computer Science & 0.1043 & 0.09596 \\
Economics & 0.1071 & 0.1073 \\
Engineering & 0.02049 & 0.02116 \\
Health & 0.2634 & 0.1301 \\
History & 0.1520 & 0.06088 \\
Law & 0.1904 & 0.2429 \\
Math & 0.1243 & 0.1368 \\
Other & 0.07913 & 0.08483 \\
Philosophy & 0.06537 & 0.04759 \\
Physics & 0.2744 & 0.2465 \\
Psychology & 0.09043 & 0.07677 \\
\midrule
Overall & 0.1444 & 0.06056 \\
\bottomrule
\end{tabular}
\end{table}

\FloatBarrier

\Needspace{0.55\textheight}
\subsection{Reconstruction and Behavioral Compositionality Across Six SAEs}

\begin{figure}[!htbp]
  \centering
  \includegraphics[width=0.98\textwidth]{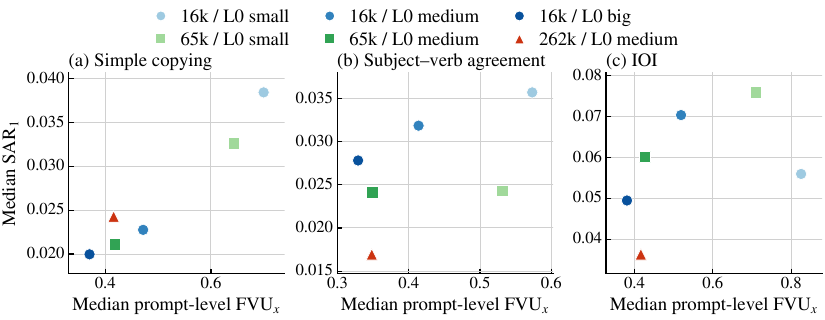}
  \caption{Median prompt level $\mathrm{FVU}_x$ and median
  $\widehat{\mathrm{SAR}}_1$
  for six SAEs at Gemma-3 1B layer 17. Panels show simple copying,
  subject verb agreement, and IOI.}
  \label{fig:reconstruction-compositionality}
\end{figure}

Figure~\ref{fig:reconstruction-compositionality} shows the comparison across the six SAE bases referenced in the main text.

\FloatBarrier

\Needspace{0.55\textheight}
\subsection{Examples}

\begin{figure}[!htbp]
  \centering
  \includegraphics[width=0.98\textwidth]{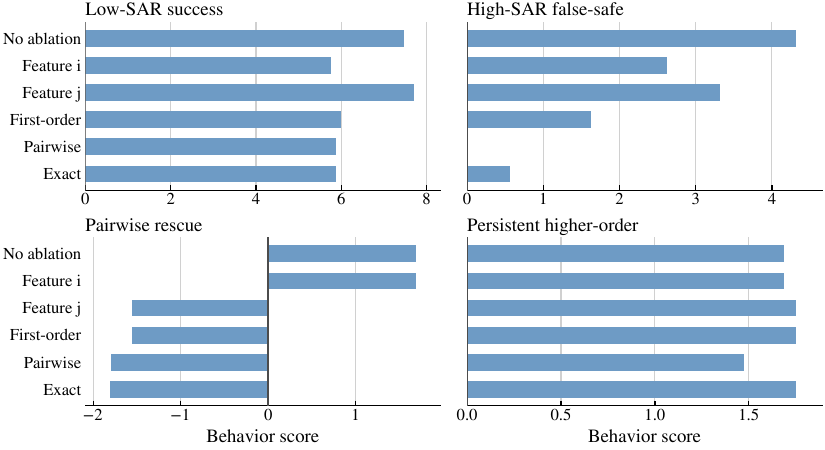}
  \caption{Behavior scores for four examples under ablations of individual
  features and feature pairs, together with first order and pairwise
  predictions when available.}
  \label{fig:case-studies}
\end{figure}

Figure~\ref{fig:case-studies} and Table~\ref{tab:case-studies} show the selected examples for individual prompts and their residual values.

\begin{table}[!htbp]
\caption{Prompts used in Figure~\ref{fig:case-studies}.}
\label{tab:case-studies}
\centering
\small
\setlength{\tabcolsep}{1.8pt}
\begin{tabular}{@{}clp{0.25\textwidth}p{0.25\textwidth}ccc@{}}
\toprule
Panel & Model & Behavior & Prompt ID & $\smash[t]{\widehat{\SAR}}_1$ & $\widehat{\SAR}_2$ & $\widehat{\rho}_2$ \\
\midrule
(a) & Mistral-7B & Simple copying & \path{simple_copying/02819} & 0.00325 & 0.00159 & 0.509 \\
(b) & Llama-3.1 8B & Gendered pronoun preference & \path{gendered_pronoun/00310} & 0.1677 & 0.0384 & 0.771 \\
(c) & Llama-3.1 8B & Subject verb agreement & \path{subject_verb_agreement/03544} & 0.00765 & 0.00103 & 0.865 \\
(d) & Llama-3.1 8B & Gendered pronoun preference & \path{gendered_pronoun/00072} & 0.1412 & 0.0417 & 0.705 \\
\bottomrule
\end{tabular}
\end{table}

\FloatBarrier

\section{Robustness and Protocol Checks}
\label{app:robustness}

\Needspace{0.48\textheight}
\subsection{Sensitivity to the number of selected features}

Changing $K$ changes the candidate explanation by adding or removing SAE
coordinates. The table therefore reports how behavioral compositionality changes
with the explanation size rather than convergence of a fixed quantity.

\begin{table}[!htbp]
\caption{Sensitivity of $\smash[t]{\widehat{\SAR}}_1$ to the number of selected SAE
features. The active feature rows report prompt median $\smash[t]{\widehat{\SAR}}_1$ at
Gemma-3 1B layer 17.
Affine control values are reported in units of $10^{-14}$. The steering row
reports $\smash[t]{\widehat{\SAR}}_1$ for the overall MMLU-Pro accuracy game. Feature sets are
nested across $K$. The sensitivity experiment uses 384 sampled coalitions
(256 for fitting, 128 for evaluation), while the main steering comparison
uses 1,024 coalitions (512 for fitting, 512 for evaluation).}
\label{tab:k-sensitivity}
\centering
\small
\setlength{\tabcolsep}{3.5pt}
\begin{tabular*}{\textwidth}{@{\extracolsep{\fill}}lcccccccc@{}}
\toprule
& \multicolumn{8}{c}{Number of selected features $K$} \\
\cmidrule(lr){2-9}
Behavior or setting
& 6 & 8 & 10 & 12 & 14 & 16 & 18 & 20 \\
\midrule
Affine control
& 6.87 & 6.40 & 5.70 & 5.16 & 5.03 & 5.43 & 5.07 & 4.84 \\
Greater than
& 0.0460 & 0.0575 & 0.0634 & 0.0707 & 0.0763 & 0.0823 & 0.0868 & 0.0872 \\
IOI
& 0.0554 & 0.0559 & 0.0597 & 0.0640 & 0.0652 & 0.0662 & 0.0664 & 0.0693 \\
Simple copying
& 0.0213 & 0.0217 & 0.0234 & 0.0245 & 0.0270 & 0.0299 & 0.0307 & 0.0320 \\
\midrule
MMLU-Pro steering
& 0.2577 & 0.4951 & 0.6595 & 0.6350 & 0.5804 & 0.7768 & 0.8096 & 0.7861 \\
\bottomrule
\end{tabular*}
\end{table}

Increasing the coalition inclusion probability from $0.5$ to $0.65$ preserves
the decrease in $\smash[t]{\widehat{\SAR}}_1$ from the earliest to the latest sampled
layer for
CounterFact, IOI, and simple copying. Mean and median aggregation give the same
depth directions. Using 1,000 prompts also preserves the main summary results,
including decreases in $\smash[t]{\widehat{\SAR}}_1$ in eight of nine comparisons and
increases in $\widehat{\rho}_2$ in all nine.

\end{document}